\documentclass[final]{cvpr}
\usepackage[T1]{fontenc}
\usepackage[latin9]{inputenc}
\usepackage{array}
\usepackage{multirow}
\usepackage{amsmath}
\usepackage{amssymb}
\usepackage{graphicx}

\makeatletter

\providecommand{\tabularnewline}{\\}

\usepackage{times}
\usepackage{epsfig}
\usepackage{graphicx}
\usepackage{amsmath}
\usepackage{amssymb}
\usepackage[boxed,algo2e]{algorithm2e}

\usepackage[linewidth=1pt]{mdframed}

\begin{document}

\title{Ordinal Pooling}
\author{
	\hspace{0.5cm}Adrien Deli\`ege\\
	\hspace{0.5cm}{\small University of Li\`ege}\\
	\and
	\hspace{0.5cm}Maxime Istasse\\
	\hspace{0.5cm}{\small University of Louvain}\\
	\and
	\hspace{0.5cm}Ashwani Kumar\\
	\hspace{0.5cm}{\small University of Sheffield}\\
	\and
	\hspace{0.5cm}Christophe De Vleeschouwer\\
	\hspace{0.5cm}{\small University of Louvain}\\
	\and
	\hspace{0.5cm}Marc Van Droogenbroeck\\
	\hspace{0.5cm}{\small University of Li\`ege}\\
}

\maketitle

\begin{mdframed}
	\textbf{How to cite this work?} This is the authors' preprint version of a paper published at BMVC 2019. Please cite it as follows: A. Deliège, M. Istasse, A. Kumar, C. De Vleeschouwer and M. Van Droogenbroeck, "Ordinal Pooling", in \textit{British Machine Vision Conference}, 2019.
\end{mdframed}

\begin{abstract}
In the framework of convolutional neural networks, downsampling is
often performed with an average-pooling, where all the activations
are treated equally, or with a max-pooling operation that only retains
an element with maximum activation while discarding the others. Both
of these operations are restrictive and have previously been shown
to be sub-optimal. To address this issue, a novel pooling scheme,
named\emph{ ordinal pooling}, is introduced in this work. Ordinal
pooling rearranges all the elements of a pooling region in a sequence
and assigns a different weight to each element based upon its order
in the sequence. These weights are used to compute the pooling operation
as a weighted sum of the rearranged elements of the pooling region.
They are learned via a standard gradient-based training, allowing
to learn a behavior anywhere in the spectrum of average-pooling to
max-pooling in a differentiable manner. Our experiments suggest that
it is advantageous for the networks to perform different types of
pooling operations within a pooling layer and that a hybrid behavior
between average- and max-pooling is often beneficial. More importantly,
they also demonstrate that ordinal pooling leads to consistent improvements
in the accuracy over average- or max-pooling operations while speeding
up the training and alleviating the issue of the choice of the pooling
operations and activation functions to be used in the networks. In
particular, ordinal pooling mainly helps on lightweight or quantized
deep learning architectures, as typically considered e.g. for embedded
applications. Code will be available at \url{https://github.com/mistasse/ordinal-pooling-layers}.
\end{abstract}

\global\long\def\comma{\enspace\mbox{,}}%

\global\long\def\dot{\enspace\mbox{.}}%


\global\long\def\convolutionSymbol{*}%


\global\long\def\weight{w}%

\global\long\def\kernel{\weight}%

\global\long\def\kernelElementOrWeight#1#2{\kernel_{#1}^{#2}}%

\global\long\def\kernelTemplate{\mathtt{w}}%

\global\long\def\kernelTemplateNumbered#1{\kernelTemplate_{\mathtt{#1}}}%

\global\long\def\half{\frac{1}{2}}%
\global\long\def\third{\text{\ensuremath{\frac{1}{3}}}}%
\global\long\def\quater{\frac{1}{4}}%
~

\section{Introduction}

Convolutional neural networks (CNNs)~\cite{Lecun1998Gradient} that
are specifically suited for various visual tasks, such as image classification~\cite{Krizhevsky2012ImageNet},
object detection~\cite{Ren2017Faster}, segmentation~\cite{He2017Mask},
and modeling video evolution~\cite{Fernando2015Modeling}, are one
of the main drivers of deep learning. A typical deep CNN architecture
consists of three types of layers: 1) convolutional: for extracting
various features or activations from an input image or feature maps,
2) pooling: a downsampling technique for aggregating elements within
a pooling region so that the size of the feature maps along the spatial
dimensions becomes smaller, and 3) fully connected: to carry out the
classification from the extracted features at the end of the network.
Many types of CNNs have been reported in the literature, for instance,
network-in-network (NIN)~\cite{Lin2013Network}, residual networks
(ResNets)~\cite{He2016DeepResidual}, inception networks~\cite{Szegedy2017Inception},
squeeze-and-excitation networks (SENets)~\cite{Hu2017Squeeze}, densely
connected convolutional networks (DenseNets)~\cite{Huang2017Densely}.

Replicating convolution kernels across the spatial dimensions in CNNs
enables weight sharing across space. This helps achieve equivariance,
i.e. a translation of an object in an input image results in an equivalent
translation in the activations of the output feature map. The pooling
operation, on the other hand, tends to achieve translational invariance,
i.e. a translation of an object in an input image does not influence
the output of the network. This pooling operation is most commonly
performed either by average-pooling (shortened to avg-pooling in the
following), where all the activations in a pooling region are averaged
together, or by max-pooling, where only the element with the maximum
activation is retained. A theoretical analysis on these two pooling
operations reveals that none of the techniques is optimal~\cite{Boureau2010ATheoretical}.
Yet, it has sometimes been argued that max-pooling achieves better
performances over avg-pooling because avg-pooling treats all the elements
equivalently irrespective of their activations, which results in an
undervaluation of the elements with higher activations, while the
elements with smaller activations are overestimated~\cite{Scherer2010Evaluation,Boureau2011Ask}.

This work presents an alternative pooling scheme, named \emph{ordinal
pooling,} that generalizes the classic avg- and max-pooling operations
and resolves the issue of unfair valuation of the elements in a pooling
region, while still preserving the information from other activations.
In this scheme, all the elements in a pooling region are first ordered
based upon their activations and then combined together via a weighted
sum, where the weights are assigned depending upon the orders of the
elements and are learned with a standard gradient-based optimization
during the training phase. Moreover, a key difference between ordinal
pooling and a classic pooling layer is that while a typical pooling
acts upon each feature map in the same way, ordinal pooling learns
a different set of weights for each feature map and therefore allows
much more flexibility in the pooling layer.

\section{Related Works}

The idea of a rank-based weighted aggregation was first introduced
by Kolesnikov \emph{et al.}~\cite{Kolesnikov2016Seed} in the context
of image segmentation, who proposed a global weighted rank-pooling
(GWRP) in order to estimate a score associated with a segmentation
class. However, GWRP is used only as a global pooling procedure as
it acts upon all the elements in a feature map to generate the score
of a particular segmentation class. Also, contrary to ordinal pooling,
the weights that are assigned based on the order of the elements are
determined from a hyper-parameter and therefore do not change during
the training.

In addition to GWRP~\cite{Kolesnikov2016Seed}, other variants of
rank-based pooling have been introduced by Shi \emph{et al.}~\cite{Shi2016Rank},
who proposed three pooling schemes based upon the rank of the elements:
1) average, 2) weighted, and 3) stochastic. Unlike ordinal pooling,
all of these schemes require an additional hyperparameter, which is
thus not learned in a differentiable way. Indeed, in the first scheme,
the hyperparameter is fixed to determine the threshold for choosing
the activations to be averaged. In the second scheme, it is fixed
to generate the weights to be applied to the activations, which remain
the same across all the feature maps, while in the third scheme, a
set of probabilities is generated based upon this hyperparameter and
is used to select an element in a pooling region.

Other works focus upon generalizing the pooling operation. Gulcehre
\emph{et al.}~\cite{Gulcehre2014Learned} regard pooling as a $l_{p}$
norm, where the values of $1$ and $\infty$ for the parameter $p$
correspond to avg- and max-pooling, whereas $p$ itself is learned
during the training. Pinheiro \emph{et al.}~\cite{Pinheiro2015FromImage}
use a smooth convex approximation of max-pooling, called Log-Sum-Exp,
where a hyperparameter controls the smoothness of the approximation,
so that pixels with similar scores have a similar weight in the training
process. Lee \emph{et al.}~\cite{Lee2018Generalizing} propose mixing
together avg-pooling and max-pooling by a trainable parameter and
also introduce the idea of tree pooling to learn different pooling
filters and combine these filters responsively.

Since a $2\times2$ max-pooling with a stride of 2 in each spatial
dimension discards $75\%$ of a feature map upon its application,
it is an aggressive operation, which after a series of applications
can result in a significant loss in information. To apply pooling
in a gentler manner, a fractional max-pooling~\cite{Graham2014Fractional}
has been proposed, where the dimensions of the feature map can be
reduced by a non-integer factor. In the spirit of allowing information
from other activations within a pooling region to also pass to the
next layer, a stochastic version of pooling has been proposed by Zeiler
\emph{et al.}~\cite{Zeiler2013Stochastic}, where an element in a
pooling region is selected based upon its probability within the multinomial
distribution constructed from all the activations inside the pooling
region. Another stochastic variant of pooling, S3Pool~\cite{Zhai2017S3Pool},
is a two-step pooling technique, where in the first step, a $2\times2$
pooling with a stride of $1$ is applied, while in the second step,
a stochastic downsampling is performed. A combination of these operations
makes S3Pool to work as a strong regularization technique.

\section{Method }

A pooling operator can be seen as a real-valued function $f_{P}$
defined on the finite non empty subsets of real numbers $\mathcal{P}_{\text{\text{fin}}}(\mathbb{R})=\{A\subset\mathbb{R}:0<|A|<+\infty\}$.
In particular, the avg-pooling and max-pooling operators, noted $f_{P}^{\text{avg}}$
and $f_{P}^{\text{max}}$, are respectively defined by
\begin{equation}
f_{P}^{\text{avg}}(A)=\frac{1}{|A|}\sum_{a\in A}a,\ \ \ \ \ \ \  f_{P}^{\text{max}}(A)=\max\{a:a\in A\}.
\end{equation}

In CNNs, a pooling layer is used to decrease the spatial resolution
of the feature maps obtained after the application of a nonlinear
activation on responses to trainable convolutional filters. A pooling
layer thus transforms an input tensor $s\in\mathbb{R}^{H\times W\times C}$
of spatial resolution $H\times W$ with $C$ feature maps (or channels)
to an output tensor $t\in\mathbb{\mathbb{R}}^{H'\times W'\times C}$
with $H'<H$ and $W'<W$. This is commonly done via a $m\times n$
pooling operation, which consists of slicing $s$ into $I$ pooling
regions $R_{i}\in\mathbb{R}^{m\times n\times C}$ ($i\leq I$), and
applying a same pooling operator $f_{P}$ to each channel $R_{i}^{c}\in\mathbb{R}^{m\times n}$
($c\leq C$) in each $R_{i}$. In a conventional CNN, the same pooling
operator $f_{P}\in\{f_{P}^{\text{avg}},f_{P}^{\text{max}}\}$ is used
for all the feature maps and remains fixed as the network trains. 

In this work, we introduce the \emph{ordinal pooling} layer, whose
pooling operator involves trainable weights that are specific to each
feature map. In the case of a $m\times n$ ordinal pooling layer,
a trainable weight kernel $\kernel^{c}\in\mathbb{\mathbb{R}}^{m\times n}$
is used to pool the regions $R_{i}^{c}\in\mathbb{R}^{m\times n}$
($i\leq I$) located on the feature map $c$ of the input tensor $s$.
The ordinal pooling operator associated with $\kernel^{c}$, defined
on $\mathbb{R}^{m\times n}$, is given by

\begin{equation}
f_{P,\kernel^{c}}^{\text{ord}}(R_{i}^{c})=\kernel^{c}\convolutionSymbol\text{Ord}(R_{i}^{c})=\sum_{j\leq m,\,k\leq n}\kernelElementOrWeight{j,k}c\text{Ord}(R_{i}^{c})_{j,k}\comma
\end{equation}
where $\text{Ord}()$ is a function from $\mathbb{R}^{m\times n}$
to $\mathbb{R}^{m\times n}$ that reorders the values of its input
tensor based upon a given ranking process. In this work, we consider
that $\text{Ord}(A)$ reorders the activations of a tensor $A\in\mathbb{R}^{m\times n}$
based upon the decreasing order of their values, such that for $j,j'\leq m$
and $k,k'\leq n$:
\begin{equation}
(j<j')\lor(k<k')\Rightarrow\text{Ord}(A)_{j,k}\geq\text{Ord}(A)_{j',k'}.
\end{equation}

\begin{figure*}
\centering{}\includegraphics[width=0.99\columnwidth]{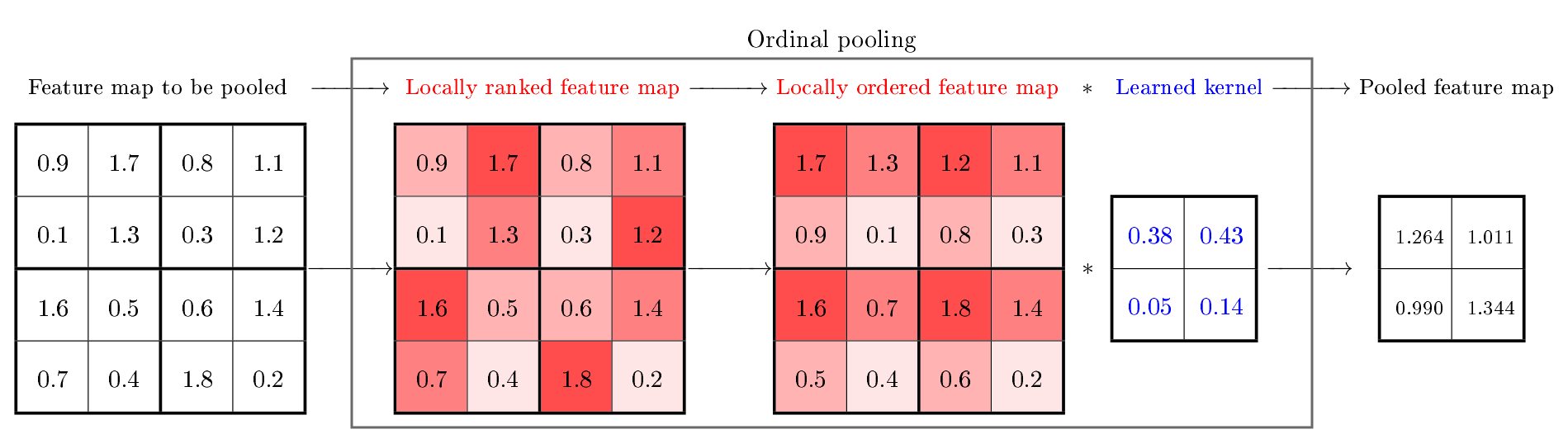}\caption{Example of $2\times2$ ordinal pooling with stride 2 in each dimension
for one feature map. The feature map to be pooled is sliced into $2\times2$
regions, whose elements are ranked by decreasing order, i.e. in each
region, the largest (resp. smallest) value has rank $1$ (resp. $4$)
and is colored in dark (resp. light) red. Then, each region is reordered
following the ranks of its elements and is convolved with the learned
kernel associated with this feature map.}
\label{fig:ordinalpooling}
\end{figure*}

This implies that, for example, $\kernelElementOrWeight{1,1}c$(resp.
$\kernelElementOrWeight{m,n}c$) always multiplies the largest (resp.
smallest) value of $R_{i}^{c}$, for all $i\leq I$ and $c\leq C$.
An illustration of $2\times2$ ordinal pooling is represented in Figure~\ref{fig:ordinalpooling}.
In practice, we constrain each kernel $w^{c}$ to contain only positive
weights that sum to $1$. This is imposed to adhere to the common
principle that a pooling operation is designed to aggregate the values
comprised in a tensor and should thus output a value located in its
convex hull. In particular, this guarantees that the output value
is comprised between the minimum and the maximum values of the input
tensor. An algorithm of the main workflow for the forward pass and
the update of the weights is detailed in supplementary material to
show how ordinal pooling can be implemented for the usual $2\times2$
case. Let us note that, since ordinal pooling employs a different
set of weights for each feature map, the total number of parameters
introduced by this operation is $m\times n\times C$, which is negligible
compared with convolutional and fully connected layers.

Ordinal pooling generalizes the commonly used avg- and max- pooling
operators. Indeed, $m\times n$ avg-pooling is a particular case of
ordinal pooling for which $\kernelElementOrWeight{j,k}c=1/(mn)$ for
all $j\leq m$, $k\leq n$. Likewise, $m\times n$ max-pooling corresponds
to the case where $\kernelElementOrWeight{1,1}c=1$ and $\kernelElementOrWeight{j,k}c=0$
for $(j,k)\neq(1,1)$. Also, compared with the other trainable pooling
operations in the literature, ordinal pooling is the only technique
that can lead to a min-pooling behavior.

\section{Experiments}

\subsection{Proof-of-concept}

\paragraph*{Setup. }

We perform the following proof-of-concept experiment on MNIST. Let
us consider a baseline network N comprising average pooling layers
and its ordinal counterpart ON, wherein the average pooling layers
are replaced by ordinal pooling layers. Since N and ON have the same
structure, we initialize them exactly in the same way with the same
weights for the non-pooling layers. The sole difference between N
and ON is the additional weights required by ordinal pooling layers.
These weights are initialized with ``average pooling'' initialization,
i.e. for an ordinal pooling kernel of size $m\times n$, each weight
is initialized as $1/(mn)$. This implies that, before starting the
training of a baseline network N and its ON counterpart, the two networks
are exactly in the same state, they produce the same output if they
are fed with the same input. Moreover, we fix all the random seeds,
so that the two networks will experience exactly the same batches
of images, in the same order, the data augmentation is the same, at
any time, over the course of their training. To guarantee the reproducibility
of the experiments and avoid suffering from GPU-based non-determinism,
these experiments are carried out on CPU. This setting allows us to
compare the results of N and ON pairwise, for each run of the experiment,
which provides a fairer and more significant insight on the intrinsic
superiority of one network over the other.

\paragraph*{Networks compared. }

Three ``baseline networks'' are used in the experiment, described
as follows with standard compact notations: 
\begin{enumerate}
\item ``Baseline'': $5\times5$ Conv $\times32$, $2\times2$ pooling,
$5\times5$ Conv $\times64$, $8\times8$ global pooling, FC($10$),
softmax.
\item ``Baseline-2'': $3\times3$ Conv $\times16$ ZP, $2\times2$ pooling,
$3\times3$ Conv $\times32$ ZP, $2\times2$ pooling, $3\times3$
Conv $\times64$ ZP, $7\times7$ global pooling, FC($10$), softmax.
\item ``LeNet5'': $5\times5$ Conv $\times6$ ZP, $2\times2$ pooling,
$5\times5$ Conv $\times16$, $2\times2$ pooling, FC($120$), ReLU,
FC($84$), ReLU, FC($10$), softmax.
\end{enumerate}
These networks have their ``ordinal'' counterpart, e.g. ``Ordinal
baseline-2'', for which the layers ``(global) pooling'' are replaced
by ``(global) ordinal pooling''. As mentioned above, ``(global)
average pooling'' are used as pooling layers in the classic pooling
setting while ``average pooling initialization'' is used to instantiate
the weights of the ordinal pooling kernels. More details about the
training of these networks are provided in supplementary material.

\paragraph*{Results. }

\begin{figure}[t]
\begin{centering}
\includegraphics[width=0.45\columnwidth]{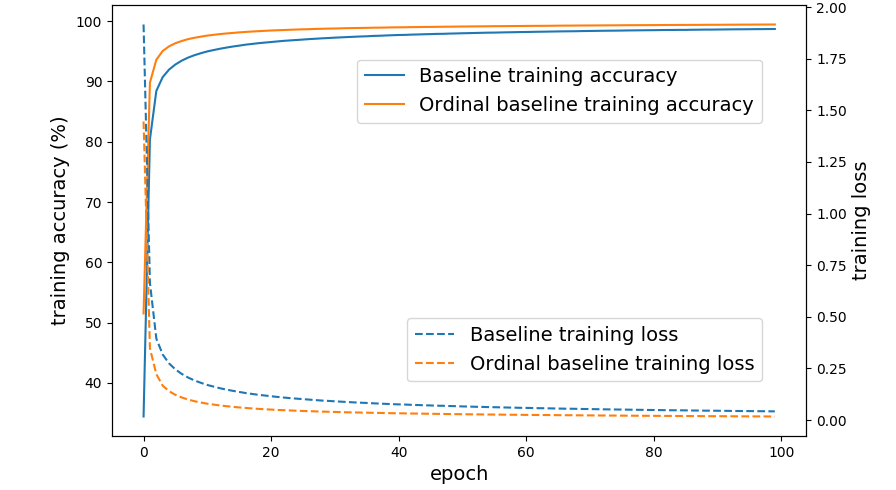}\includegraphics[width=0.45\columnwidth]{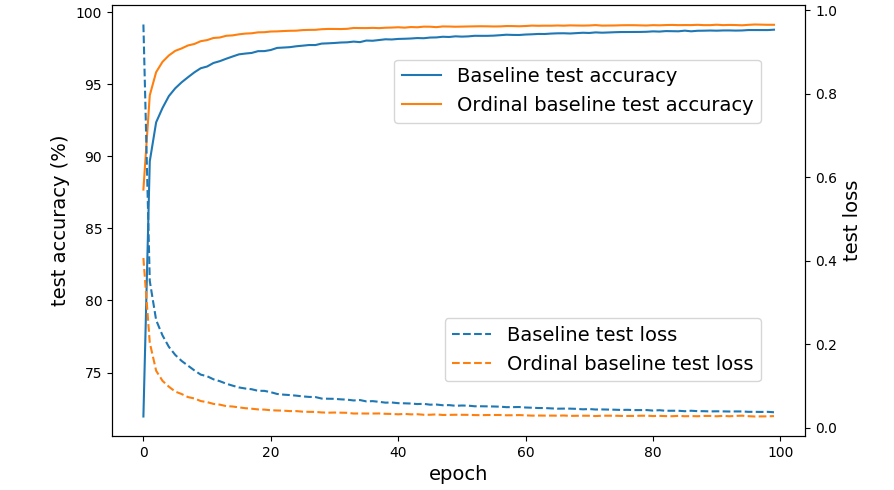}
\par\end{centering}
\centering{}\includegraphics[width=0.45\columnwidth]{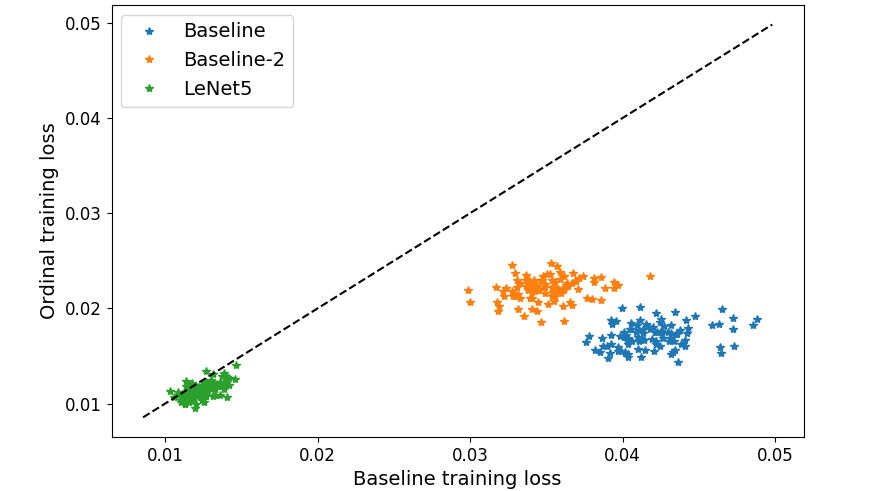}\includegraphics[width=0.45\columnwidth]{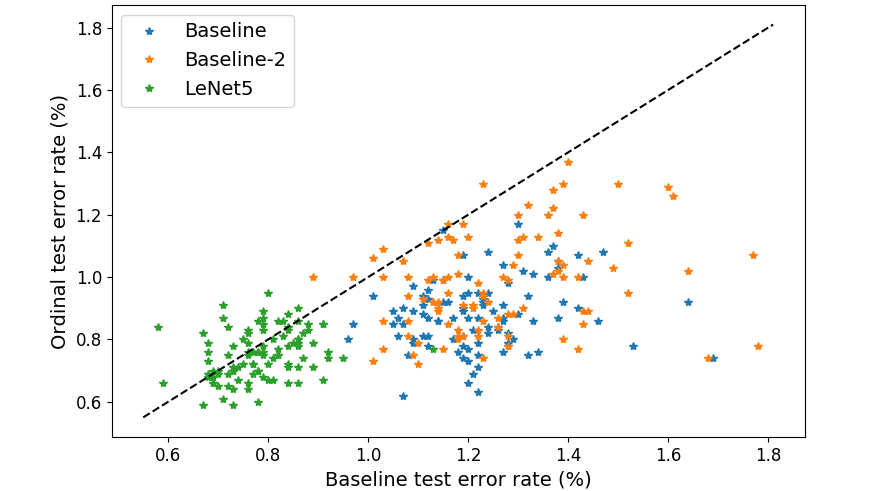}\caption{Average learning curves for the ``Baseline'' network and its ``Ordinal
baseline'' counterpart and pairwise comparison of different runs
(differing in random seeds initializations) of the networks with and
without ordinal pooling having the same instantiation. Points under
the black dotted line indicate better performances for networks with
ordinal pooling. }
\label{fig:poc-learning-curves}
\end{figure}

Each network is run $100$ times, where the runs differ by their initial
random seeds. Figure~\ref{fig:poc-learning-curves} shows the average
learning curves for the network ``Baseline'' and its ``Ordinal
baseline'' counterpart. It can be seen that ordinal pooling allows
to reach better performances in terms of accuracy and loss, while
it also speeds up the training process.

The pairwise comparison of the performances of the networks versus
their ordinal counterpart is represented in Figure~\ref{fig:poc-learning-curves}.
As can be seen, the ``Baseline'' and ``Baseline-2'' networks employing
ordinal pooling always achieve smaller training loss at the end of
the training. The same is also true for the ``LeNet5'' network $89\%$
of the time. Regarding test error rates, the ordinal pooling networks
outperform the classic ones $100\%$, $94\%$, and $74\%$ of the
time for the ``Baseline'', ``Baseline-2'', and ``LeNet5'' cases,
respectively. Even though ordinal pooling seems less beneficial to
LeNet5, the results have to be put in perspective with respect to
the extra cost in parameters that the ordinal pooling layers require.
In fact, as can be inferred from Table~\ref{tab:relatives}, LeNet5
has by far the best ratio between the relative improvement in performances
(both in terms of average and variance in test error rate) and the
number of additional parameters contained in the ordinal pooling layers.
Table~\ref{tab:relatives} suggests that overall, the performances
of the networks are boosted with ordinal pooling by a comfortable
margin through a reduced average test error rate, and that ordinal
pooling provides a more consistent convergence between different runs,
through a reduced variance in test error rate. These benefits come
at a moderate cost in terms of number of parameters.

\begin{table}
\begin{centering}
{\small{}}%
\begin{tabular}{|c|c|c|c|}
\hline 
{\small{}Relative variation in} & {\small{}Baseline} & {\small{}Baseline-2} & {\small{}LeNet5}\tabularnewline
\hline 
\hline 
{\small{}average/variance of training loss} & {\small{}$-59\%/-70\%$} & {\small{}$-37\%/-66\%$} & {\small{}$-8\%/-25\%$}\tabularnewline
\hline 
{\small{}average/variance of test loss} & {\small{}$-25\%/-5\%$} & {\small{}$-21\%/-8\%$} & {\small{}$-5\%/-23\%$}\tabularnewline
\hline 
{\small{}average/variance of test error rate} & {\small{}$-28\%/-26\%$} & {\small{}$-22\%/-18\%$} & {\small{}$-4\%/-7\%$}\tabularnewline
\hline 
{\small{}number of parameters} & {\small{}$+8\%$} & {\small{}$+14\%$} & {\small{}$+0.14\%$}\tabularnewline
\hline 
\end{tabular}{\small\par}
\par\end{centering}
\hspace{1cm}
\centering{}\caption{Relative variation in some metrics when the avg-pooling layers of
the baseline networks are replaced by ordinal pooling layers. Networks
with ordinal pooling show a large decrease in the average and variance
of the losses and error rates for a moderate increase in number of
parameters.}
\label{tab:relatives}
\end{table}

\paragraph*{{\scriptsize{}}}

\paragraph*{Distribution of $2\times2$ ordinal pooling kernels. }

The use of ordinal pooling instead of a classic pooling operation
allows to study the distributions of the weight kernels in the ordinal
pooling layers, as learned by the networks, and helps discover how
the trained networks chose to perform the pooling operations. For
that purpose, we compare the learned kernels with some \emph{template
kernels} that characterize various categories of behaviors for the
kernels, including avg- and max-pooling like behaviors\emph{. }These
template kernels are chosen based on the behavior that they induce
as explained below. 

In the case of a $2\times2$ ordinal pooling layer, a weight kernel
$\kernel=\left[\kernelElementOrWeight 1{},\kernelElementOrWeight 2{},\kernelElementOrWeight 3{},\kernelElementOrWeight 4{}\right]$
leads to max-pooling if it converges to $[1,0,0,0]$ and to avg-pooling
if each $w_{i}=1/4$. In the first case, the network ``promotes''
only the largest value of each pooling region, while in the second
case, all the values are equally ``promoted''. Nevertheless, a network
may prefer to promote, for example, the lowest value of the regions
(thus min-pooling behavior) by using the kernel $\left[0,0,0,1\right]$,
or its first two largest values with $\left[1/2,1/2,0,0\right]$.
The template kernels are determined based upon this idea of enumerating
all the ways that some values can be promoted by the network. In
fact, it has the possibility to promote any of the four ordered values
of the regions by making $w$ converge to one of the following four
template kernels:{\scriptsize{}
\[
\kernelTemplateNumbered 1=\left[1,0,0,0\right],\ \kernelTemplateNumbered 2=\left[0,1,0,0\right],\ \kernelTemplateNumbered 3=\left[0,0,1,0\right],\ \kernelTemplateNumbered 4=\left[0,0,0,1\right].
\]
}In the same spirit, it may prefer to promote equally two, three,
or the four values of the ordered regions, making $w$ converge to{\scriptsize{}
\begin{align*}
\kernelTemplateNumbered{12} & =\left[\half,\half,0,0\right],\ \kernelTemplateNumbered{13}=\left[\half,0,\half,0\right],\ \kernelTemplateNumbered{14}=\left[\half,0,0,\half\right],\ \kernelTemplateNumbered{23}=\left[0,\half,\half,0\right],\ \kernelTemplateNumbered{24}=\left[0,\half,0,\half\right],\ \kernelTemplateNumbered{34}=\left[0,0,\half,\half\right],\\
 & \kernelTemplateNumbered{123}=\left[\third,\third,\third,0\right],\ \kernelTemplateNumbered{124}=\left[\third,\third,0,\third\right],\kernelTemplateNumbered{134}=\left[\third,0,\third,\third\right],\ \kernelTemplateNumbered{234}=\left[0,\third,\third,\third\right],\ \kernelTemplateNumbered{1234}=\left[\quater,\quater,\quater,\quater\right].
\end{align*}
}{\scriptsize\par}

\begin{figure}
\begin{centering}
\includegraphics[width=0.5\columnwidth]{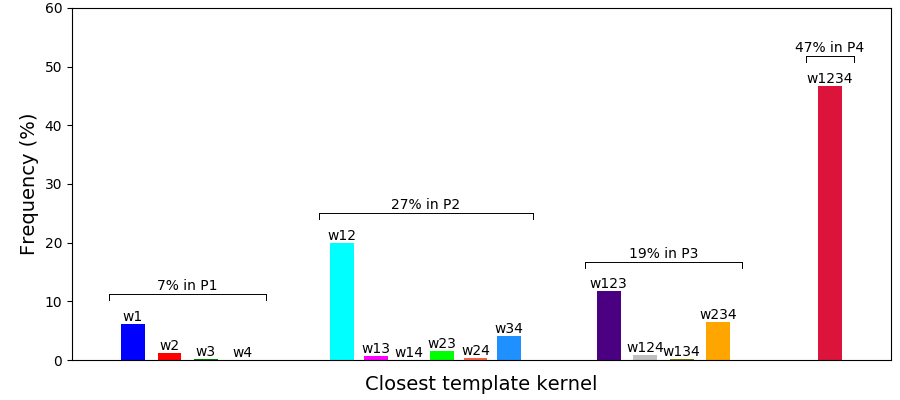}\hfill{}\includegraphics[width=0.5\columnwidth]{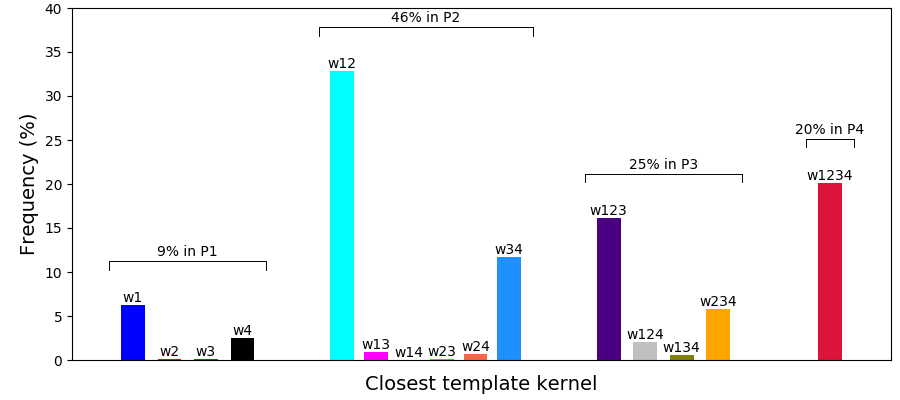}
\par\end{centering}
\centering{}\includegraphics[width=0.5\columnwidth]{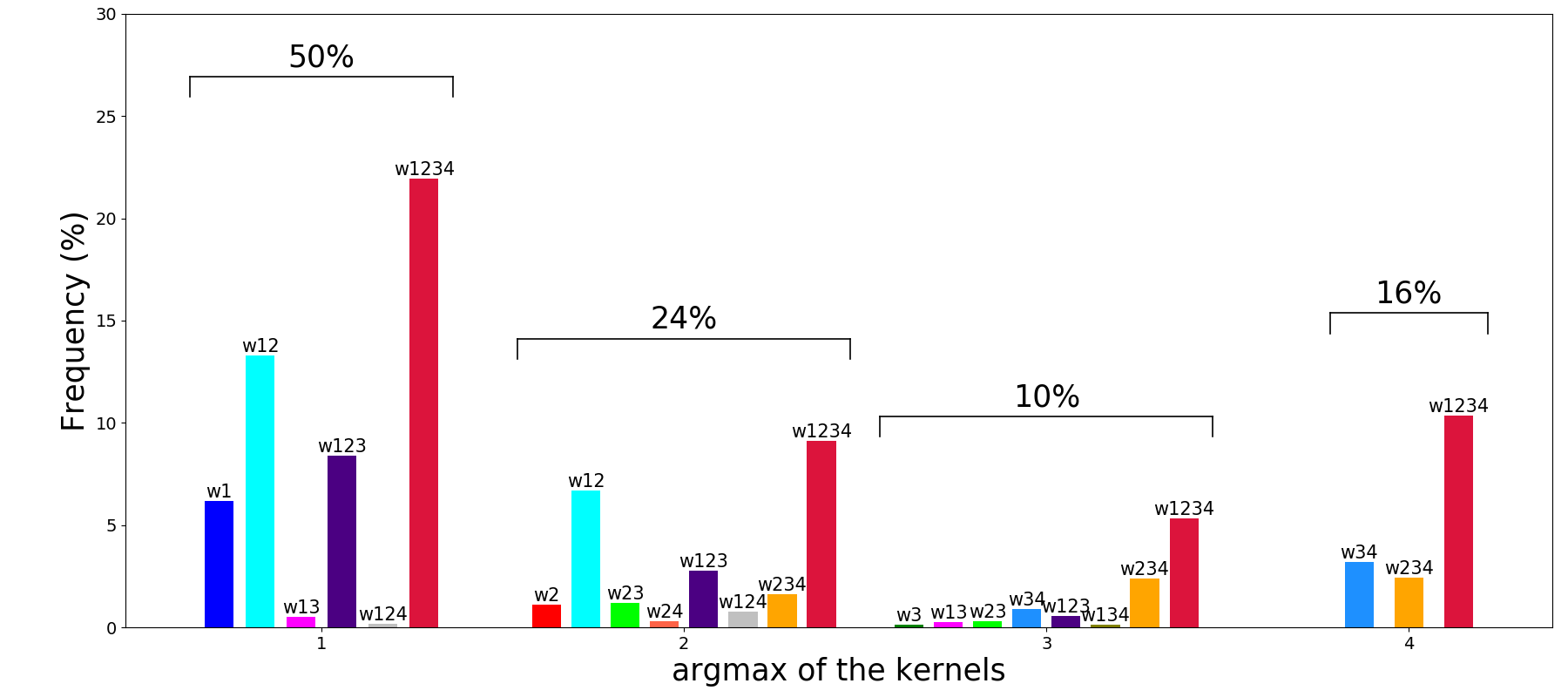}\hfill{}\includegraphics[width=0.5\columnwidth]{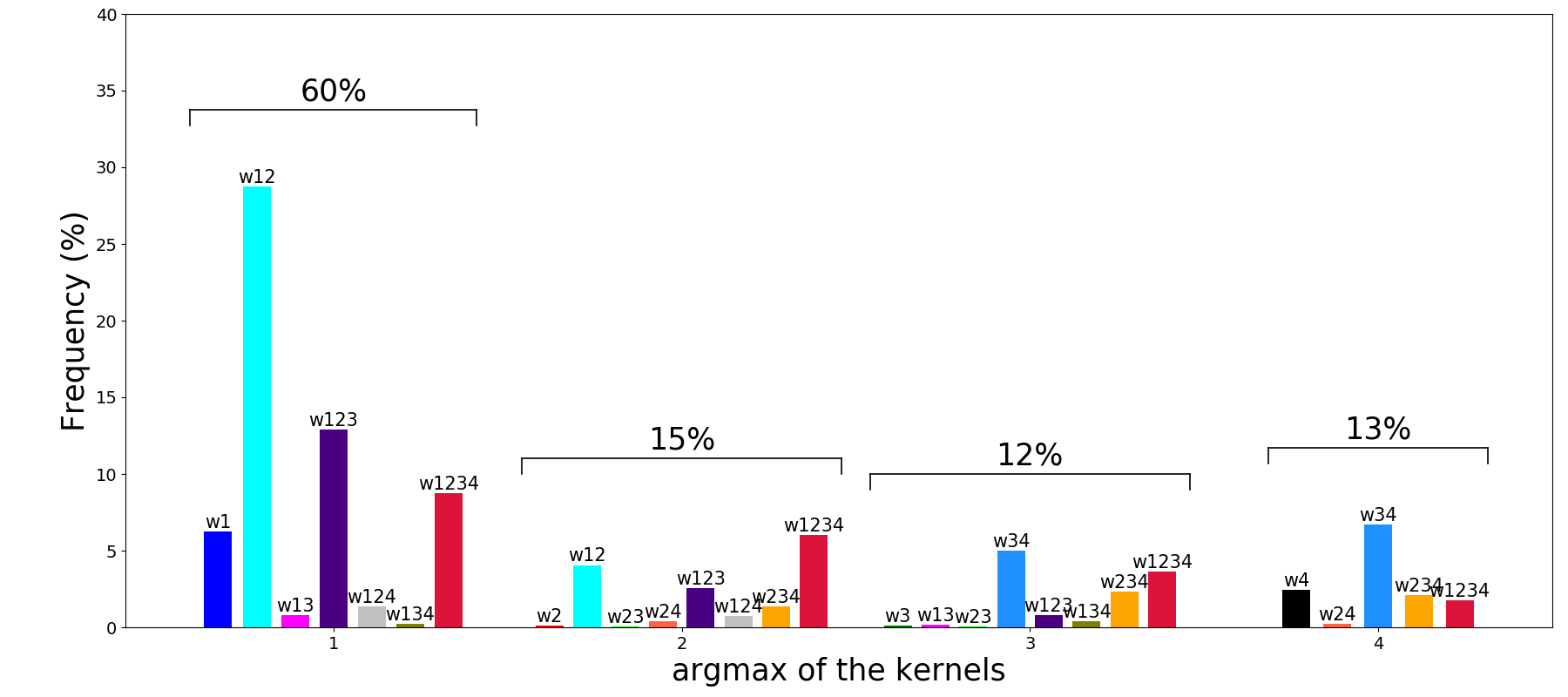}\caption{Distributions of the kernels of the first (left) and second (right)
ordinal pooling layers of ``Ordinal baseline-2'' according to their
closest template kernel, grouped by number of promoted values (top),
and grouped by the index of their largest weight (bottom).}
\label{fig:distributions-groups}
\end{figure}

We note $P_{i}$ the set of template kernels having $i$ non-zero
values. After the training of a network, for each kernel of a $2\times2$
ordinal pooling layer, we identify its closest template kernel, in
term of Euclidean distance. We examine the distribution of the learned
kernels by grouping them by ``closest template kernels'' to find
out how the network chooses to perform the pooling operations. These
distributions for ordinal pooling layers of ``Baseline-2'', grouped
by $P_{i}$ and aggregated for the $100$ runs, are displayed in Figure~\ref{fig:distributions-groups}
(top left and top right).

It can be seen that, even though all the kernels were initialized
as $\kernelTemplateNumbered{1234},$ after the training less than
half of them remain closer to $\kernelTemplateNumbered{1234}$ than
to any other template kernel. Another observation is that the network
seems to prefer promoting contiguous ``extreme'' (largest or smallest)
values in the sorted regions, i.e. when $2$ (resp. $3$) values are
promoted, the associated kernels are preferably closer to $\kernelTemplateNumbered{12}$
or $\kernelTemplateNumbered{34}$ (resp. $\kernelTemplateNumbered{123}$
or $\kernelTemplateNumbered{234}$). The ``extreme'' aspect is reinforced
in the group $P_{1}$ of the top right plot, in which almost only
$\kernelTemplate_{1}$ and $\kernelTemplate_{4}$ are present. Hence,
some kernels actually display a ``min-pooling'' behavior.

Also, the network prefers to promote the largest value of the regions,
which manifests by the fact that the $\text{argmax}$ of a learned
kernel is often $1$, thus indicating that a behavior between average
pooling and max-pooling is often desired. This observation is illustrated
in Figure~\ref{fig:distributions-groups} (bottom left and bottom
right), where the kernels are first distributed according to their
$\text{argmax}$, then sub-divided following their closest template
kernel. Similar trends are also observed for ``Ordinal baseline''
and ``Ordinal LeNet5'' (provided in supplementary material). A
complementary analysis of the kernels related to the global ordinal
pooling layer is presented in supplementary material.

\subsection{Influence of ordinal weights initialization}

In this section, we examine the performances of the networks under
various weight initializations in the ordinal pooling layers. As
before, we report average test error rates over $100$ runs of each
experiment to compare the results in Table~\ref{tab:performances-weightinit}.

For the $2\times2$ ordinal pooling kernels, the initializations investigated
are average, max, min. For the average (resp. max, min) case, each
kernel is instantiated as $\kernelTemplateNumbered{1234}$ (resp.
$\kernelTemplateNumbered 1$,$\kernelTemplateNumbered 4$). For the
global $n\times n$ ordinal pooling layers, average pooling ($1/n^{2}$)
is used.

Table~\ref{tab:performances-weightinit} shows that the networks
with ordinal pooling consistently outperform the classic ones. Also,
the performances are less dependent on the initialization of the ordinal
pooling kernels than they are in the classic setting. Ordinal pooling
thus alleviates the problem of choosing the appropriate type of pooling
layer to incorporate in the networks. 

Max-pooling initialization performs better in this experiment. However,
even with min-pooling initialization, the ordinal networks are still
able to reach performances close to standard initializations, while
it is not as clear for the classic networks. An explanation may reside
in the use of ReLU activations before the pooling layers. In the classic
setting, min-pooling forces the networks to select the lowest value
of the pooled regions, hence it can be assumed that many zero values
are propagated in the network, responsible for decreasing the amount
of useful information and thus leading to lower performances. In the
ordinal pooling setting, the network has enough flexibility to circumvent
this fixed min-pooling behavior, and this only requires small variations
in the ordinal weights. Indeed, we observed that the closest template
kernel after the training was still most often $\kernelTemplate_{4}$.

We also examined a ``uniform'' initialization, where for each kernel,
all the weights are randomly initialized with uniform distribution
between $0$ and $1$ and are then normalized so as to sum to $1$.
The results do not differ much from those obtained with ``average''
initialization and are discussed in supplementary material.

\begin{table}
\begin{centering}
{\small{}}{\small\par}
\par\end{centering}
\begin{centering}
{\small{}}%
\begin{tabular}{|c|c|c|c|c|c|c|c|c|c|}
\cline{1-6} \cline{2-6} \cline{3-6} \cline{4-6} \cline{5-6} \cline{6-6} \cline{8-10} \cline{9-10} \cline{10-10} 
\multirow{2}{*}{{\small{}Pooling}} & \multirow{2}{*}{{\small{}$2\times2$}} & \multirow{2}{*}{{\small{}glob. }} & \multirow{2}{*}{{\small{}Bas.}} & \multirow{2}{*}{{\small{}Bas.-2}} & \multirow{2}{*}{{\small{}LN5}} &  & \multicolumn{3}{c|}{{\small{}Activation Bas.-2}}\tabularnewline
\cline{8-10} \cline{9-10} \cline{10-10} 
 &  &  &  &  &  &  & {\small{}None} & {\small{}ReLU} & {\small{}tanh}\tabularnewline
\cline{1-6} \cline{2-6} \cline{3-6} \cline{4-6} \cline{5-6} \cline{6-6} \cline{8-10} \cline{9-10} \cline{10-10} 
{\small{}classic} & {\small{}average} & {\small{}average} & {\small{}1.22} & {\small{}1.26} & {\small{}0.79} &  & {\small{}49.50} & {\small{}1.26} & {\small{}1.62}\tabularnewline
\cline{1-6} \cline{2-6} \cline{3-6} \cline{4-6} \cline{5-6} \cline{6-6} \cline{8-10} \cline{9-10} \cline{10-10} 
{\small{}ordinal} & {\small{}average} & {\small{}average} & {\small{}0.89} & {\small{}1.00} & {\small{}0.75} &  & {\small{}1.13} & {\small{}1.00} & {\small{}0.99}\tabularnewline
\cline{1-6} \cline{2-6} \cline{3-6} \cline{4-6} \cline{5-6} \cline{6-6} \cline{8-10} \cline{9-10} \cline{10-10} 
{\small{}classic} & {\small{}max} & {\small{}average} & {\small{}0.98} & {\small{}1.01} & {\small{}0.74} &  & {\small{}3.69} & {\small{}1.01} & {\small{}1.58}\tabularnewline
\cline{1-6} \cline{2-6} \cline{3-6} \cline{4-6} \cline{5-6} \cline{6-6} \cline{8-10} \cline{9-10} \cline{10-10} 
{\small{}ordinal} & {\small{}max} & {\small{}average} & {\small{}0.82} & {\small{}0.89} & {\small{}0.71} &  & {\small{}1.01} & {\small{}0.89} & {\small{}0.90}\tabularnewline
\cline{1-6} \cline{2-6} \cline{3-6} \cline{4-6} \cline{5-6} \cline{6-6} \cline{8-10} \cline{9-10} \cline{10-10} 
{\small{}classic} & {\small{}min} & {\small{}average} & {\small{}1.48} & {\small{}1.34} & {\small{}0.94} &  & {\small{}3.82} & {\small{}1.34} & {\small{}1.55}\tabularnewline
\cline{1-6} \cline{2-6} \cline{3-6} \cline{4-6} \cline{5-6} \cline{6-6} \cline{8-10} \cline{9-10} \cline{10-10} 
{\small{}ordinal} & {\small{}min} & {\small{}average} & {\small{}0.97} & {\small{}1.02} & {\small{}0.84} &  & {\small{}1.05} & {\small{}1.02} & {\small{}0.90}\tabularnewline
\cline{1-6} \cline{2-6} \cline{3-6} \cline{4-6} \cline{5-6} \cline{6-6} \cline{8-10} \cline{9-10} \cline{10-10} 
\end{tabular}{\small\par}
\par\end{centering}
\centering{}\hspace{1cm}\caption{(Left) Average test error rates (in $\%$) for various weight initializations
in the pooling layers. For classic pooling, the weights are not trainable.
(Right) Corresponding average test error rates for ``Baseline-2''
and its ordinal counterpart with various activation functions.}
\label{tab:performances-weightinit}
\end{table}

\subsection{Influence of activation functions}

The case of the ordinal min-pooling initialization raises the question
of the influence of the activation function used before the pooling
operation. We thus compare the average test error rates of various
initializations and three types of activation functions: ``None''
(no activation), ``ReLU'', ``tanh''. The results for the ``Baseline-2''
structure are reported in Table~\ref{tab:performances-weightinit}.

The results obtained for a given initialization with ordinal pooling
are less sensitive to the choice of the activation function than those
obtained in the networks with classic pooling schemes. Conversely,
for a given activation, the results with ordinal pooling are less
sensitive to the choice of initialization compared to the networks
employing classic poolings. 

One of the most striking results may be related to the performances
obtained without any activation. Indeed, while it is well-known that
CNNs need non-linear activations to achieve competitive performances,
the networks with ordinal pooling layers still manage to obtain good
performances without activation. In this case, some are even better
than others obtained in the classic setting with activations. The
sorting procedure in the ordinal pooling layer is itself a non-linearity,
which explains these results. This is especially true for the avg-pooling
case, where it is known that an avg-pooling layer without prior activation
is useless. For better performances, it still appears that using an
activation is beneficial even with ordinal pooling layers, but the
choice of the function may not be as crucial as in the networks with
classic pooling. Similar trends were observed with ``Baseline''
and ``LeNet5'' structures.

\subsection{Results on other datasets and best use cases}

Ordinal pooling can be used in any CNN architecture involving pooling
layers. Its benefits vary from one use case to another, as indicated
by the following additional results, reported as average test error
rates on five trials. On CIFAR10, with a CNN made of five Conv(128)-ReLU-Pooling
blocks and a FC layer, ordinal pooling outperforms avg-pooling ($13.16\%$
vs $14.21\%$). With DenseNet-BC-100-12~\cite{Huang2017Densely},
the results are mostly equivalent with ordinal and avg-pooling ($5.53\%$
vs $5.45\%$ with ReLU, $6.35\%$ vs $6.85\%$ with tanh) except without
activation ($12.51\%$ vs $59.39\%$), similarly to CIFAR100 ($24.73\%$
vs $24.74\%$ with ReLU, $27.01\%$ vs $27.05\%$ with tanh, $37.39\%$
vs $81.61\%$ without activation). 

Even though exhaustive performance-related experiments still need
to be carried out as future work, the present results are in line
with those reported previously and confirm that ordinal pooling mainly
helps on relatively simple architectures, as typically considered
e.g. for embedded applications. To further support this statement,
we performed experiments with quantized networks, as described in~\cite{Moons2017Minimum}.
It appears that the more the model is quantized, the more ordinal
pooling helps: with quantized versions of ResNet-14 (resp. ResNet-20),
on CIFAR10, ordinal pooling performs up to $3.5
$ (resp.~$1.1\%$) better than max-pooling. It also reduces the gap
between binary ResNet-14 and -20, which is of $0.7\%$, against $2.7\%$
with max-pooling, which certainly opens interesting prospects for
ordinal pooling, as it helps simpler models achieve performances comparable
with more complex models.

For the record, our experiments on MNIST and CIFAR10 with \cite{Lee2018Generalizing}
lead to results comparable with those presented above with ordinal
pooling ($<1\%$ difference). A comprehensive comparison with the
pooling methods present in the literature could be the subject of
a survey article (along with defining benchmark tests to assess the
performances of a pooling method), and is thus beyond the scope of
this work.

\section{Conclusion}

A novel trainable pooling scheme, \emph{Ordinal Pooling}, is introduced
in this work, which operates in two steps. In the first step, all
the elements of a pooling region are reordered in decreasing sequence.
Then, a trainable weight kernel is convolved with the rearranged pooling
region to compute the output of the ordinal pooling operation. The
usual avg- and max-pooling operations can be recovered as particular
cases of ordinal pooling. 

In our experiments, replacing classic avg- and max-pooling operations
with ordinal pooling produces large relative improvements in classification
performances at a moderate cost in additional parameters and also
leads to a faster convergence. Ordinal pooling allows to perform the
pooling operation differently in distinct feature maps. The analysis
of the learned kernels reveals that the networks take advantage of
this extra flexibility by using various types of pooling for different
feature maps within the same pooling layer. A general trend is that
a hybrid behavior between avg- and max-pooling is often desired, even
though the lowest elements of the pooling regions are not always discarded.
Moreover, the performances of the networks are less inclined to fluctuate
when different initializations of the ordinal pooling kernels are
used than when different classic pooling operations are imposed. Besides,
even when no non-linear activation function is applied after the convolutional
layers, the intrinsic non-linearity introduced by ordinal pooling
alone generally suffices to produce performances which are better
than either avg- and max-pooling used along with activation functions.
Finally, our experiments suggest that ordinal pooling might be of
particular interest for lightweight or quantized architectures, as
typically considered in e.g. embedded resource-constrained systems. 

As future work, as already mentioned, it will be interesting to perform
more experiments with more datasets and various architectures to determine
the configurations which best benefit from the ordinal pooling operation.
From a technical point of view, the value of the elements of the pooling
regions is chosen as the criterion for ordering the region. However,
other criteria could also be envisioned to further extend the ordinal
pooling scheme. Eventually, conducting experiments with our baseline
models but with other types of pooling methods proposed in the literature
is certainly our next step in order to rank ordinal pooling among
available pooling operations.

\paragraph*{Acknowledgements}

This research is supported by the DeepSport project of the Walloon
region, Belgium, C. De Vleeschouwer is funded by the F.R.S.-FNRS.


\begin{thebibliography}{10}\itemsep=-1pt
	
	\bibitem{Boureau2011Ask}
	Y.~Boureau, N.~{Le Roux}, F.~Bach, J.~Ponce, and Y.~Lecun.
	\newblock Ask the locals: Multi-way local pooling for image recognition.
	\newblock In {\em IEEE Int. Conf. Comput. Vision (ICCV)}, pages 2651--2658,
	Barcelona, Spain, Nov. 2011.
	
	\bibitem{Boureau2010ATheoretical}
	Y.~Boureau, J.~Ponce, and Y.~LeCun.
	\newblock A theoretical analysis of feature pooling in visual recognition.
	\newblock In {\em Int. Conf. Mach. Learn. (ICML)}, pages 111--118, Haifa,
	Israel, June 2010.
	
	\bibitem{Fernando2015Modeling}
	B.~Fernando, E.~Gavves, M.~{Jose Oramas}, A.~Ghodrati, and T.~Tuytelaars.
	\newblock Modeling video evolution for action recognition.
	\newblock In {\em IEEE Int. Conf. Comput. Vision and Pattern Recogn. (CVPR)},
	pages 5378--5387, Boston, MA, USA, June 2015.
	
	\bibitem{Graham2014Fractional}
	B.~Graham.
	\newblock Fractional max-pooling.
	\newblock {\em CoRR}, abs/1412.6071, Dec. 2014.
	
	\bibitem{Gulcehre2014Learned}
	C.~Gulcehre, K.~Cho, R.~Pascanu, and Y.~Bengio.
	\newblock Learned-norm pooling for deep feedforward and recurrent neural
	networks.
	\newblock In {\em Machine Learning and Knowledge Discovery in Databases},
	volume 8724 of {\em Lecture Notes Comp. Sci.}, pages 530--546. Springer,
	2014.
	
	\bibitem{He2017Mask}
	K.~He, G.~Gkioxari, P.~Dollar, and R.~Girshick.
	\newblock Mask {R-CNN}.
	\newblock In {\em IEEE Int. Conf. Comput. Vision (ICCV)}, pages 2980--2988,
	Venice, Italy, Oct. 2017.
	
	\bibitem{He2016DeepResidual}
	K.~He, X.~Zhang, S.~Ren, and J.~Sun.
	\newblock Deep residual learning for image recognition.
	\newblock In {\em IEEE Int. Conf. Comput. Vision and Pattern Recogn. (CVPR)},
	pages 770--778, Las Vegas, NV, USA, June 2016.
	
	\bibitem{Hu2017Squeeze}
	J.~Hu, L.~Shen, and G.~Sun.
	\newblock Squeeze-and-excitation networks.
	\newblock {\em CoRR}, abs/1709.01507, 2017.
	
	\bibitem{Huang2017Densely}
	G.~Huang, Z.~Liu, L.~{van der Maaten}, and K.~Weinberger.
	\newblock Densely connected convolutional networks.
	\newblock In {\em IEEE Int. Conf. Comput. Vision and Pattern Recogn. (CVPR)},
	pages 2261--2269, Honolulu, HI, USA, July 2017.
	
	\bibitem{Kolesnikov2016Seed}
	A.~Kolesnikov and C.~Lampert.
	\newblock Seed, expand and constrain: Three principles for weakly-supervised
	image segmentation.
	\newblock In {\em Eur. Conf. Comput. Vision (ECCV)}, volume 9908 of {\em
		Lecture Notes Comp. Sci.}, pages 695--711. Springer, 2016.
	
	\bibitem{Krizhevsky2012ImageNet}
	A.~Krizhevsky, I.~Sutskever, and G.~Hinton.
	\newblock Image{N}et classification with deep convolutional neural networks.
	\newblock In {\em Adv. in Neural Inform. Process. Syst. (NeurIPS)}, volume~25,
	pages 1097--1105, 2012.
	
	\bibitem{Lecun1998Gradient}
	Y.~Lecun, L.~Bottou, Y.~Bengio, and P.~Haffner.
	\newblock Gradient-based learning applied to document recognition.
	\newblock {\em Proc. of IEEE}, 86(11):2278--2324, Nov. 1998.
	
	\bibitem{Lee2018Generalizing}
	C.-Y. Lee, P.~Gallagher, and Z.~Tu.
	\newblock Generalizing pooling functions in {CNN}s: Mixed, gated, and tree.
	\newblock {\em IEEE Trans. Pattern Anal. Mach. Intell.}, 40(4):863--875, Apr.
	2018.
	
	\bibitem{Lin2013Network}
	M.~Lin, Q.~Chen, and S.~Yan.
	\newblock Network in network.
	\newblock {\em CoRR}, abs/1312.4400, Dec. 2013.
	
	\bibitem{Moons2017Minimum}
	B.~Moons, K.~Goetschalckx, N.~{Van Berckelaer}, and M.~Verhelst.
	\newblock Minimum energy quantized neural networks.
	\newblock In {\em Asilomar Conference on Signals, Systems, and Computers},
	pages 1921--1925, Pacific Grove, CA, USA, 2017.
	
	\bibitem{Pinheiro2015FromImage}
	P.~O. Pinheiro and R.~Collobert.
	\newblock From image-level to pixel-level labeling with convolutional networks.
	\newblock In {\em IEEE Int. Conf. Comput. Vision and Pattern Recogn. (CVPR)},
	jun 2015.
	
	\bibitem{Ren2017Faster}
	S.~Ren, K.~He, R.~Girshick, and J.~Sun.
	\newblock {Faster R-CNN}: Towards real-time object detection with region
	proposal networks.
	\newblock {\em IEEE Trans. Pattern Anal. Mach. Intell.}, 39(6):1137--1149, June
	2017.
	
	\bibitem{Scherer2010Evaluation}
	D.~Scherer, A.~M{\"{u}}ller, and S.~Behnke.
	\newblock Evaluation of pooling operations in convolutional architectures for
	object recognition.
	\newblock In {\em Int. Conf. Artificial Neural Networks (ICANN)}, volume 6354
	of {\em Lecture Notes Comp. Sci.}, pages 92--101. Springer, 2010.
	
	\bibitem{Shi2016Rank}
	Z.~Shi, Y.~Ye, and Y.~Wu.
	\newblock Rank-based pooling for deep convolutional neural networks.
	\newblock {\em Neural Networks}, 83:21--31, Nov. 2016.
	
	\bibitem{Szegedy2017Inception}
	C.~Szegedy, S.~Ioffe, V.~Vanhoucke, and A.~Alemi.
	\newblock Inception-v4, {I}nception-{ResNet} and the impact of residual
	connections on learning.
	\newblock In {\em AAAI Conf. Artificial Intell.}, pages 4278--4284, San
	Francisco, CA, USA, Feb. 2017.
	
	\bibitem{Zeiler2013Stochastic}
	M.~Zeiler and R.~Fergus.
	\newblock Stochastic pooling for regularization of deep convolutional neural
	networks.
	\newblock In {\em Int. Conf. on Learn. Rep. (ICLR)}, Scottsdale, Arizona, May
	2013.
	
	\bibitem{Zhai2017S3Pool}
	S.~Zhai, H.~Wu, A.~Kumar, Y.~Cheng, Y.~Lu, Z.~Zhang, and R.~Feris.
	\newblock {S3Pool}: Pooling with stochastic spatial sampling.
	\newblock In {\em IEEE Int. Conf. Comput. Vision and Pattern Recogn. (CVPR)},
	pages 4003--4011, Honolulu, HI, USA, July 2017.
	
\end{thebibliography}
\end{document}